\documentclass{article}
\usepackage{spconf,amsmath,graphicx}
\usepackage{amssymb}
\usepackage{multirow}
\usepackage{subfigure}
\usepackage{multicol}
\usepackage{array}
\usepackage{float}
\usepackage{hyperref}

\usepackage{threeparttable}

\def\bW{{\mathbf W}}

\def\bs{{\mathbf s}}

\def\b1{{\mathbf 1}}

\newcommand{\etal}[1]{{#1 \textit{et al.}}}

\title{Sparse-GAN: Sparsity-constrained Generative Adversarial Network for Anomaly Detection in Retinal OCT Image}
\name{Kang Zhou$^{1,2}$\thanks{$^{1}$ {\{zhoukang, gaoshh\}@shanghaitech.edu.cn}, $^{2}$ {chengjun@nimte.ac.cn}}, Shenghua Gao$^{1,\dagger}$, Jun Cheng$^{2,3,\dagger}$\thanks{$^{\dagger}$ corresponding authors}, Zaiwang Gu$^4$, Huazhu Fu$^5$,  Zhi Tu$^1$ }
\address{
	\textit{Jianlong Yang$^2$, Yitian Zhao$^2$, Jiang Liu$^{2,4}$} \\
	$^{1}$ School of Information Science and Technology, ShanghaiTech University 
	\protect\\ $^{2}$ Cixi Institute of Biomedical Engineering, Chinese Academy of Sciences
	\protect\\ $^{3}$ UBTech Research
	\protect\\ $^4$ Southern University of Science and Technology 
	\protect\\ $^{5}$ Inception Institute of Artificial Intelligence
}

\begin{document}
%

\maketitle 

\begin{abstract}
With the development of convolutional neural network, deep learning has shown its success for retinal disease detection from optical coherence tomography (OCT) images. However, deep learning often relies on large scale labelled data for training, which is oftentimes challenging especially for disease with low occurrence. Moreover, a deep learning system trained from data-set with one or a few diseases is unable to detect other unseen diseases, which limits the practical usage of the system in disease screening. 
To address the limitation, we propose a novel anomaly detection framework termed Sparsity-constrained Generative Adversarial Network (Sparse-GAN) for disease screening where only healthy data are available in the training set.
The contributions of Sparse-GAN are two-folds: 1) The proposed Sparse-GAN predicts the anomalies in latent space rather than image-level; 2) Sparse-GAN is constrained by a novel Sparsity Regularization Net.
Furthermore, in light of the role of lesions for disease screening, we present to leverage on an anomaly activation map to show the heatmap of lesions.
We evaluate our proposed Sparse-GAN on a publicly available dataset, and the results show that the proposed method outperforms the state-of-the-art methods.
\end{abstract}

%

\vspace{-0.2cm}	
\section{Introduction}
\vspace{-0.3cm}	

\label{introduction}
Over 300 million people worldwide are affected by various ocular diseases \cite{apostolopoulos2017pathological}, such as diabetic retinopathy (DR) \cite{zhao2018uniqueness}, age-related macular degeneration (AMD) and glaucoma.
Among the many diagnostic methods, optical coherence tomography (OCT) is a non-invasive  imaging modality that provides micrometer-resolution volumetric scans of the retina \cite{huang1991optical}.
With the development of convolutional neural networks (CNNs) in computer vision \cite{krizhevsky2012imagenet, lian2018multiview}, many deep  learning based approaches have been proposed to detect lesions in retinal OCT images  \cite{lee2017deep} and fundus images \cite{zhou2018multi, gu2019net}.
However, these deep learning based methods rely heavily on big data for training, which limits the application of deep learning to medical image analysis.

\begin{figure}[ttt]
	\centering
	\includegraphics[width=3.2in]{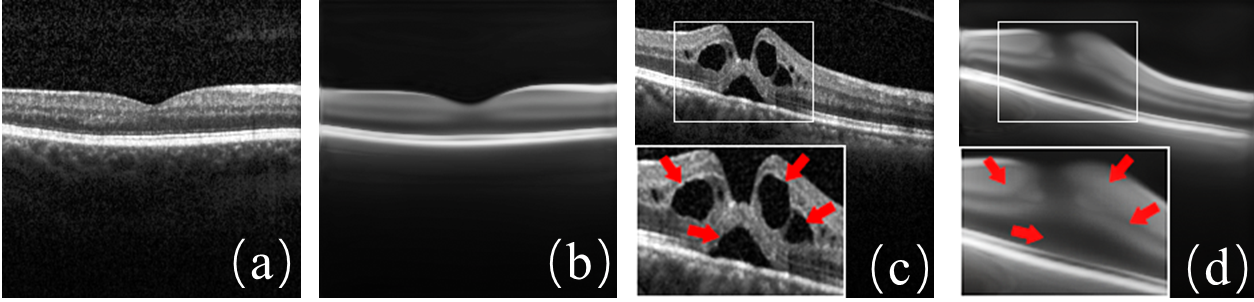}
	\caption{The input image and its reconstructed image. (a) Normal input. (b) Reconstruction of the normal input. (c) Disease input. (d) Reconstruction of the disease input with our proposed method. Since lesions can't be reconstructed, the reconstruction error is high to be recognized as abnormal.
	} \label{fig:normal_abnormal}
\vspace{-0.6cm}	
\end{figure}

Different from that in the general computer vision, it is often challenging to get sufficient data for medical images due to several reasons. The first reason is that most of the medical data is not publicly available due to privacy concerns. The second reason is that labeling medical images often costs much time, while experienced clinicians are short of time for such tedious demarcation tasks. The third reason is that the occurrence of some lesions is usually low, while the presence of specific lesions is not known before the diagnosis. Therefore, the cost of obtaining large-scale medical data with particular types of lesions is often expensive and time-consuming.

Although it is difficult to get a large amount of data with different lesions, it is often much easier to get data from healthy subjects. In OCT imaging, one 3D scan from a healthy subject could provide hundreds of B-scan images without lesions. Considering the lesions as anomaly added to the images from healthy subjects, it is possible to train an anomaly detection system only using OCT B-scans without lesions.

\begin{figure*}[ttt]
	\centering
	\includegraphics[width=5.5in]{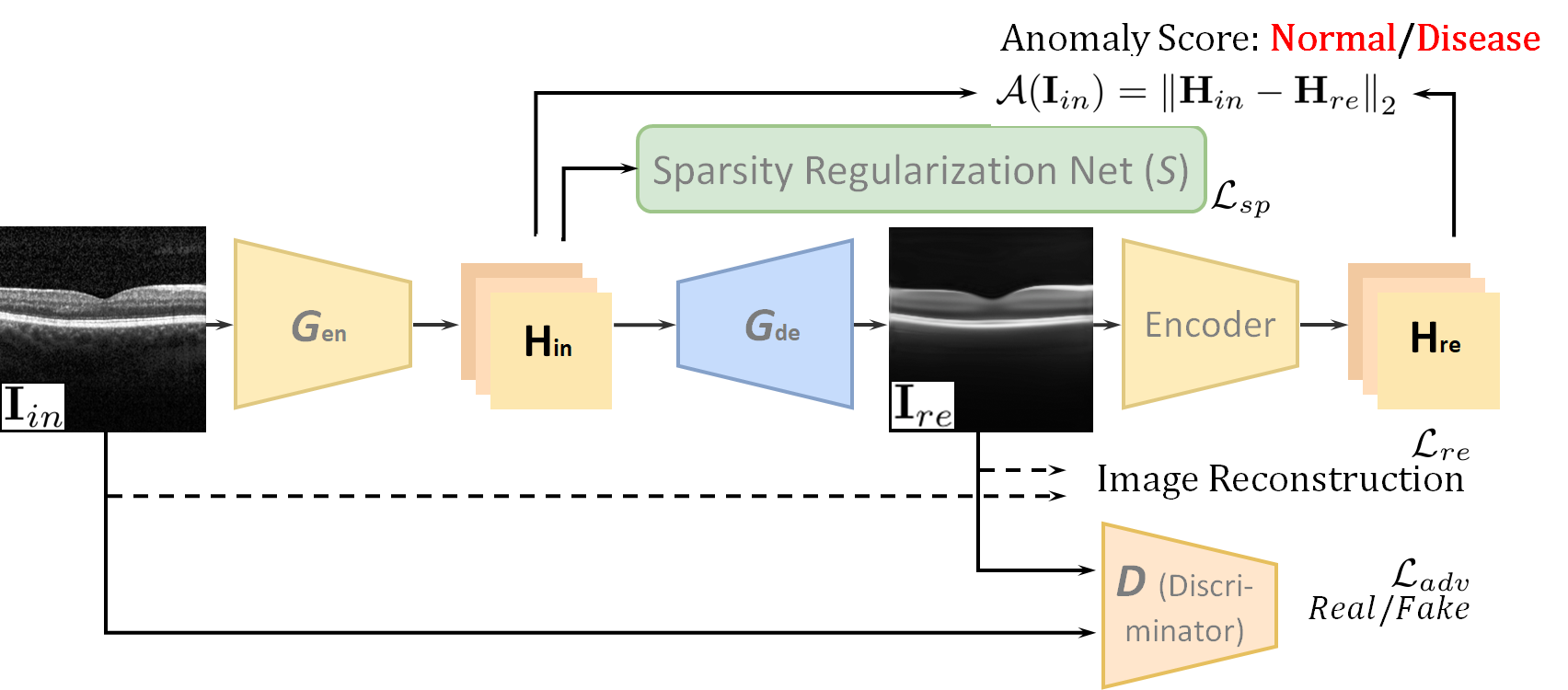}
	\caption{The overall architecture of our Sparse-GAN. Components with boxes with solid line are networks while other boxes are features. 
		In the \textbf{testing stage}, given a test image $\mathbf{I}_{in}$, firstly the image is converted into latent feature with $\mathbf{H}_{in} = G_{en}(\mathbf{I}_{in})$, while $\mathbf{H}_{in}$ is converted into reconstructed image with $\mathbf{I}_{re}= G_{de}(\mathbf{H}_{in})$. Then $\mathbf{I}_{re}$ is transformed to latent feature with another encoder $\mathbf{H}_{re} = E(\mathbf{I}_{re})$, finally the framework predicts anomaly score with $\mathcal{A}(\mathbf{I}_{in}) = \left\| \mathbf{H}_{in} - \mathbf{H}_{re} \right\|_2$. In the \textbf{training stage}, besides the same pipeline of testing, the framework is trained with image reconstruction loss $\mathcal{L}_{re}$, adversarial loss $\mathcal{L}_{adv}$ and sparsity regularization $\mathcal{L}_{sp} = S(\mathbf{H}_{in})$.	
		(Best viewed with colors.)
	\vspace{-0.6cm}		
	} \label{fig:overall_arch}
\end{figure*}

Previous work has shown the effectiveness of anomaly detection for disease diagnosis \cite{sidibe2017anomaly} and lesion location \cite{seebock2018unsupervised}.
Recently, CNNs based methods have been proposed to detect anomalies in medical images.
\etal{Schlegl} \cite{schlegl2017unsupervised} initially introduce a deep convolutional Generative Adversarial Network (GAN) \cite{goodfellow2014generative}, which is referred to a AnoGAN, to detect anomalies in OCT B-scans. Later, they further propose a f-AnoGAN \cite{schlegl2019f}, which is faster than AnoGAN.
However, these networks are not trained in an end-to-end fashion, which may tend to get stuck into local optima. It is desirable to customize a network that learns the optimal features for anomaly detection.

In this paper, inspired by Image-to-Image GAN \cite{isola2017image}, whose generator is end-to-end optimized, we propose to employ Image-to-Image GAN for medical image anomaly detection. Then, to alleviate the effect of image noise (e.g. speckle noise in OCT images), we propose to map the reconstructed image into latent space with an additional encoder. 
Furthermore, motivated by the capability of interpretable sparse coding for anomaly detection, we propose to regularize the sparsity of latent features. By taking these factors into consideration, we present a novel framework: Sparsity-constrained Generative Adversarial Network (Sparse-GAN) for image anomaly detection with merely normal training data. The rationale behind the work is that the normal patterns from healthy subjects can be reconstructed with small errors while the patterns with lesions from diseased subjects are often reconstructed with large errors, as shown in Fig. \ref{fig:normal_abnormal}.

The main contributions of this work are summarized as follows:
(1) We propose to map the images into a latent space and regularize the latent feature with a novel sparsity regularizer; (2) We introduce a novel Sparse-GAN for anomaly detection, and our method is designed for the scenario where only data corresponding to healthy subjects are available in the training set. Thus our solution may ease the difficulty in data collection and annotation; (3) Our method also predicts anomaly activation maps to show lesions for clinical diagnosis.

\vspace{-0.5cm}	
\section{Method}
\vspace{-0.3cm}	

\label{method}

In this work, we mainly focus on regularizing the sparsity of latent feature and utilizing the latent feature to predict anomalies in GAN based anomaly detection framework.
As shown in Fig. \ref{fig:overall_arch}, the proposed Sparsity-constrained Generative Adversarial Network consists of three modules: 1) Image-to-Image GAN \cite{isola2017image} for medical anomaly detection whose generator is end-to-end optimized. 2) Anomalies computing in latent space \cite{akcay2018ganomaly}, to alleviate the effect of image noise (e.g. speckle noise in OCT images). 3) The novel Sparsity Regularization Net to regularize the sparsity of latent features.

\vspace{-0.3cm}
\subsection{Image-to-Image GAN for Anomaly Detection}
\vspace{-0.3cm}

As discussed earlier, we adopt the image-to-image \cite{isola2017image} generator as the $G$ in the GAN, which consists of encoder $G_{en}$ and decoder $G_{de}$, while $D$ denotes the discriminator. Let $\mathbf{I}_{in}$ be input images, their latent feature $\mathbf{H}_{in}$ are converted from input images $\mathbf{H}_{in} = G_{en}(\mathbf{I}_{in})$, then the latent feature are transformed into reconstructed images $\mathbf{I}_{re}= G_{de}(\mathbf{H}_{in})$. Image-to-Image GAN \cite{isola2017image} is optimized with a reconstruction loss comprised of an adversarial loss, 

\begin{equation}
\min\limits_G \max\limits_D \mathcal{L}_G = \min\limits_G \left( \lambda_{adv} \max\limits_D(\mathcal{L}_{adv}) + \lambda_{re} \mathcal{L}_{re} \right),
\end{equation}
where $\lambda_{adv}$ and $\lambda_{re}$ are regularization parameters. The adversarial loss and reconstruction loss are defined as,

\begin{equation}
\mathcal{L}_{adv} = \mathbb{E}_{\mathbf{I}_{in}}[\log D(\mathbf{I}_{in})]	 + \mathbb{E}_{\mathbf{I}_{in}, \mathbf{H}_{in}}[\log(1 - D(G(\mathbf{I}_{in}), \mathbf{H}_{in}))], 
\end{equation}

\begin{equation}
\mathcal{L}_{re} = \frac{1}{m}\sum_{i=1}^m(\mathbf{I}_{in}^{(i)} - \mathbf{I}_{re}^{(i)})^2,
\end{equation}	
where $m$ is the batch-size.


\vspace{-0.2cm}
\subsection{Predict Anomaly Score in Latent Space}

One challenge in reconstructing the OCT images is the speckle noise. To reduce the influence of speckle noise, we propose to transform the reconstruction image $\mathbf{I}_{re}$ into latent space by encoder $E$, i.e.  $\mathbf{H}_{re} = E(\mathbf{I}_{re})$. To cut down computational cost, encoder $E$ share the same values with $G_{en}$. In latent space, the model predicts anomaly score $\mathcal{A}(\mathbf{I}_{in})$ and diagnosis results $\mathcal{C}(\mathbf{I}_{in})$ as follows:

\begin{equation}
\label{eq:anomaly}
\mathcal{A}(\mathbf{I}_{in}) = \left\| \mathbf{H}_{in} - \mathbf{H}_{re} \right\|_2 
= \left\| G_{en}(\mathbf{I}_{in}) - E(G(\mathbf{I}_{in}))  \right\|_2 ,
\end{equation}


\begin{equation}
\label{eq:classifcation}
\text{and  }\mathcal{C}(\mathbf{I}_{in}) = 
\begin{cases}
\text{normal,} \quad \text{if} \ \mathcal{A}(\mathbf{I}_{in}) < \phi \\
\text{disease,} \quad \text{if} \ \mathcal{A}(\mathbf{I}_{in}) \geqslant \phi
\end{cases}
\end{equation}
where $\phi$ is the anomaly score threshold determined on the validation set.

\subsection{Sparse Regularization on Latent Feature}

On the one hand, without additional regularization, generator $G$ may learn an approximation to the identity function, which can not distinguish disease images from normal images. On the other hand,  sparse coding is interpretable and have the capability for anomaly detection \cite{luo2017revisit, luo2019video}.

Based on this observation, we propose a novel Sparsity Regularization Net which recast the solution of sparse coding as a novel convolutional long short term memory unit (LSTM). Moreover, we regularize the sparsity of latent feature $\mathbf{H}_{in}$ with the proposed Sparsity Regularization Net (i.e., $S(\cdot)$) as shown in Fig. \ref{fig:overall_arch}. Letting $S$ denote Sparsity Regularization Net, we propose a novel Sparsity-constrained GAN (Sparse-GAN) with sparsity regularization $\mathcal{L}_{sp} = S(\mathbf{H}_{in})$.

The proposed Sparsity Regularization Net is inspired from Sparse LSTM \cite{zhou2018sc2net}. However, sparsity reguliarzaiton net is different from sparse LSTM in two aspects. Firstly we apply the convolutional operation to replace element-wise multiplication in Sparse LSTM since the convolutional operation accelerates the computation. Secondly the input of the Sparse Constrained Net is the latent feature rather than the original image.


The loss to train Sparsity Regularization Net is defined as follows, 
\begin{eqnarray}
\mathcal{L}_{scl}(\bW_d, \bs) = \left\| \mathbf{H}_{in} - \bW_d^T \bs \right\|_F^2 +  \left\| \bs \right \|_1
\end{eqnarray}
where $\bs$ is the sparse code w.r.t. $\mathbf{H}_{in}$ and $\bW_d$ is the dictionary.



Overall, the final loss of Sparse-GAN is given as the following:

\begin{equation}
\mathcal{L} = \lambda_{re}\mathcal{L}_{re} + \lambda_{adv} \max\limits_D(\mathcal{L}_{adv}) + \lambda_{sp}\mathcal{L}_{sp},
\end{equation}
where $\lambda_{re}, \lambda_{adv} \text{ and } \lambda_{sp}$ are regularization parameters.

\subsection{Anomaly Activation Map for Visualization}


Since anomaly detection is significantly different from supervised classification, Class Activation Map (CAM) \cite{zhou2016learning} is not suitable in our framework to show the role of lesions for diagnosis. To address the weakness of CAM, we propose Anomaly Activation Map (AAM) to visualize lesions in anomaly detection framework.
We firstly perform Global Average Pooling ($GAP$) for latent feature $\mathbf{H}_{in} \in \mathbb{R}^{1024 \times 7 \times 7}$ and $\mathbf{H}_{re} \in \mathbb{R}^{1024 \times 7 \times 7}$. Then we obtain the anomaly vector $\mathbf{W}_{aam} = w_1, w_2, \cdots, w_n$ as follows,
\begin{equation}
\mathbf{W}_{aam} = \left \| GAP(\mathbf{H}_{in}) - GAP(\mathbf{H}_{re}) \right \|_1,
\end{equation}	
where $\mathbf{W}_{aam} \in \mathbb{R}^{1024 \times 1 \times 1}$, $n$ is the number of the channels of the  latent feature. 
Finally, we multiply the feature map $\mathbf{H}_{in}$ by anomaly vector in channel-wise fashion and get the anomaly activation map.

\vspace{-0.3cm}
\section{Experiments}
\vspace{-0.3cm}

\label{experiments}

\subsection{Datasets and Evaluation Metrics}
\subsubsection{Datasets}
\vspace{-0.2cm}	

We employ a publicly available dataset \cite{kermany2018identifying} to evaluate the performance of our Sparse-GAN. The whole dataset was from Spectralis OCT (Heidelberg Engineering, German), and contains data with three different lesions: drusen, DME (diabetic macular edema), and CNV (choroidal neovascularization).
The detailed description about this dataset could be found in \cite{kermany2018identifying}. 
To train the proposed Sparse-GAN and determine the threshold of anomaly score, we divide original training set into two parts: new training set with 50,140 normal images, validation set consists of 3000 disease images and 1000 normal images. The testing set is the  same as the  original dataset.



\vspace{-0.3cm}	
\subsubsection{Evaluation Metrics}
\vspace{-0.2cm}	
For a given test image $\mathbf{I}_{in}$, we use $\mathcal{A}(\mathbf{I}_{in})$ given in Eq. (\ref{eq:anomaly}) to compute the anomaly score. Further, we use $\mathcal{C}(\mathbf{I}_{in})$ given in Eq. (\ref{eq:classifcation}) for diagnosis.
Based on the anomaly score, we mainly use \textbf{AUC} (Area under the ROC Curve) to evaluate our method. 
To compute accuracy (\textbf{Acc}), we need to  determine the threshold $\phi$ of anomaly score on the  validation set, which includes 75\% disease images and 25\% normal images. 
We adopt sensitivity (\textbf{Sen}) as the third evaluation metric.
Finally, the threshold $\phi$ is then used for testing.


\vspace{-0.3cm}
\subsection{Training Details}
The proposed Sparse-GAN is implemented in PyTorch with NVIDIA graphics processing units (GeForce TITAN V). The input image size is $224 \times 224$, while the batch size is 32. The optimizer is Adam and the learning rate is 0.001. Empirically, we let $\lambda_{re}=20, \lambda_{adv}=1$, and $\lambda_{sp}=50$. 

\vspace{-0.3cm}	
\subsection{Quantitative Experimental Results}

\vspace{-0.6cm}
\begin{table}[htb]
	\normalsize
	\setlength{\tabcolsep}{3mm}
	\centering
	\caption{Quantitative results for ablation studies and comparison  with state-of-the-arts. }
	\begin{tabular}{c|c|ccc}
		\hline
		\multirow{2}{*}{Method} & Val-set & \multicolumn{3}{c}{Test-set} \\ \cline{2-5} 
		& AUC & AUC & Acc  & Sen \\ \hline
		Auto-Encoder & 0.729 & 0.783 & 0.751 & 0.834  \\
		AnoGAN\cite{schlegl2017unsupervised} & 0.815 & 0.846 & 0.789 &  0.917  \\ 
		f-AnoGAN\cite{schlegl2019f} & 0.849 & 0.882 & 0.808 & 0.871   \\ \hline
		pix2pix \cite{isola2017image} \#1 & 0.805 & 0.861 & 0.818  & 0.879  \\
		pix2pix \cite{isola2017image} \#2 & 0.837 & 0.874 & 0.815  & 0.900  \\
		Sparse-GAN & \textbf{0.885} & \textbf{0.925} & \textbf{0.841} & \textbf{0.951} \\ \hline
	\end{tabular}

	\vspace{-0.2cm}
	\begin{flushleft}
	\small \#1, image level \\
	\small \#2, latent space \\
	\end{flushleft}
	
	\vspace{-0.3cm}
	\label{tabel:result}
\end{table}

\vspace{-1.0cm}
\subsubsection{Ablation Study.}
\vspace{-0.2cm}	
To justify the benefits of the anomaly score in latent space and the sparsity regulirization nets, we conduct the following ablation studies, we conduct  some ablation studies: \#1 denotes Image-to-Image GAN \cite{isola2017image} predicting anomaly score in image-level, and \#2 denotes Image-to-Image GAN \cite{isola2017image} predicting anomaly score  $\mathcal{A}(\mathbf{I}_{in})$ in latent feature. 

By including $\mathcal{L}_{adv}$ loss based on Auto-Encoder, we improve the AUC result from 0.729 to 0.805 on the validation set.  That is to say, adversarial learning is helpful. By transforming the reconstruction image into latent space, the result is improved from 0.805 to 0.837 on the validation set since the noise in images is harmful to diagnosis. Finally, by regularizing the latent features with our proposed Sparsity Regularization Net, the result is improved from 0.837 to 0.885, which means the sparsity regularization is effective. On the test set, the ablation studies validate the effectiveness of different modules too.
Table \ref{tabel:result} summarized the results. 

\vspace{-0.2cm}
\subsubsection{Performance Comparison.}
\vspace{-0.2cm}	
We further compare the proposed method with state-of-the-art networks, inlcuding Auto-Encoder, AnoGAN \cite{schlegl2017unsupervised} and f-AnoGAN \cite{schlegl2019f}.

By comparing our adopted Image-to-Image GAN (i.e. $\#$ 1) with primary AnoGAN \cite{schlegl2017unsupervised}, we improve the AUC result from 0.846 to 0.861 on the test set. That is to say, the end-to-end optimized generator is better than two stage trained  generator. 
Compared with these methods, we get the highest  AUC than others on both the validation set and test set. The accuracy of our method on  the test set is comparable to supervised deep learning methods, and the sensitivity $=0.951$ denotes missed diagnosis of our model is very low, which is more meaningful for clinicians. The results are also summarized in Table \ref{tabel:result} .

\vspace{-0.3cm}	
\subsection{Qualitative Analysis with Anomaly Activation Map}
\begin{figure}[ttt]
	\centering
	\includegraphics[width=3.2in]{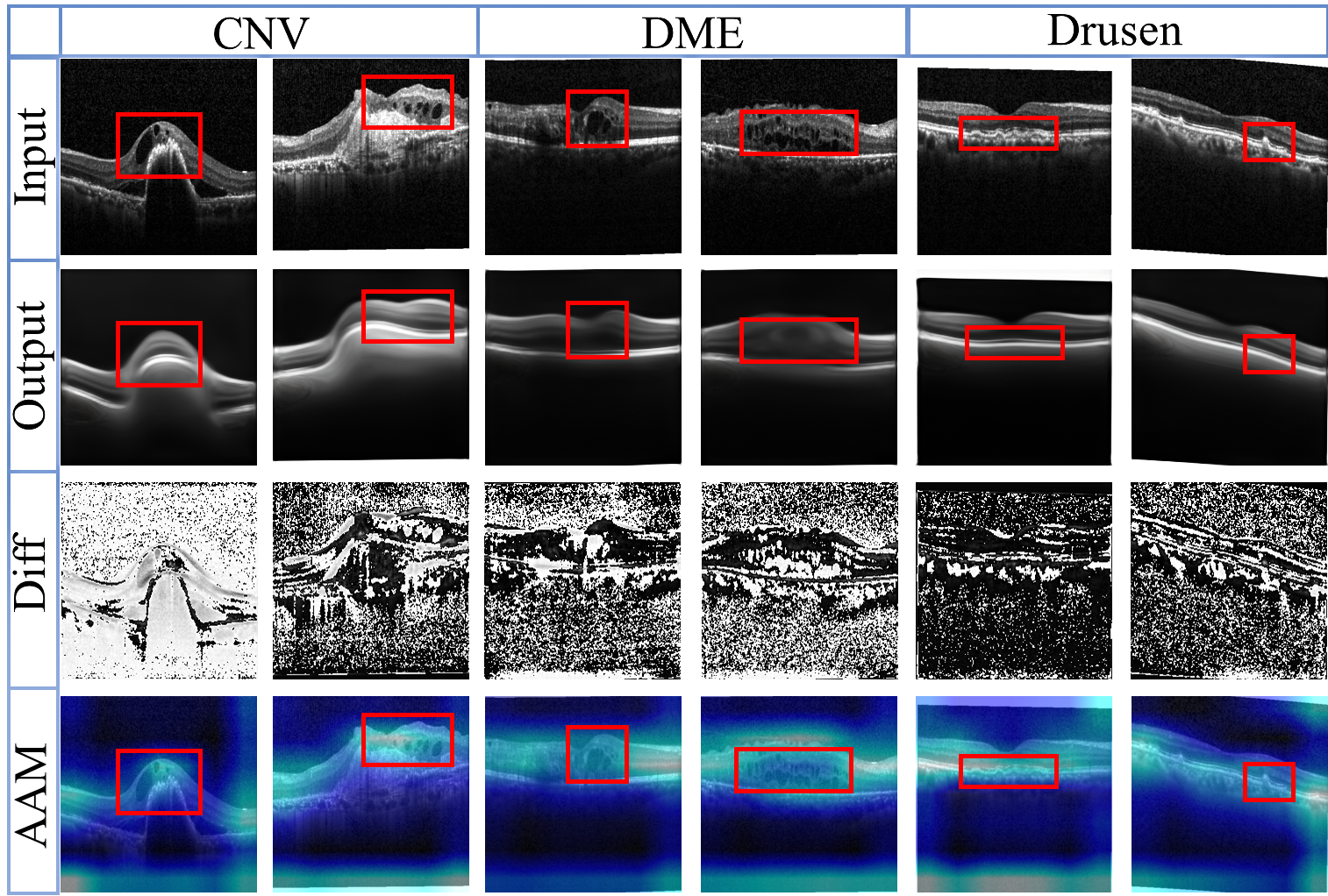}
	\caption{Anomaly heatmap on abnormal images. \textit{Diff} images show that noise in images is harmful for reconstruction, and AAM images show the lesion play an important role for diagnosis in Sparse-GAN. (Best viewed with colors.)
	} \label{fig:abnormal}
	\vspace{-0.4cm}	
\end{figure}

To further understand what the role of the lesion is for disease  clinical diagnosis, some example images are shown in Fig \ref{fig:abnormal}. When Sparse-GAN classifies a given image as abnormal, AAM will be computed. In addition to the anomaly heatmap, we also show the output images and difference between the  input image and output one.
Since Sparse-GAN is only trained on the normal set, the model could not reconstruct  abnormal patterns. \textit{Diff} images show that noise in images is harmful to  reconstruction. 
The heatmap can localize the lesion in general and this validates the effectiveness of our proposed AAM for anomaly detection framwork.

%
%


\vspace{-0.2cm}
\section{CONCLUSION}
\vspace{-0.2cm}

\label{Coucusion}
In this work, we propose a novel Sparse-GAN for anomaly detection, which detects anomalies in latent space and the feature in latent space is constrained by a novel Sparsity Regularizer Net.
The quantitative experimental results on a public dataset validate the feasibility of anomaly detection for OCT images and also validate the effectiveness of our method. Further, we also show the anomaly activation maps of the lesion to make our results more explainable.

\vspace{-0.2cm}
\section{Acknowledge}
\vspace{-0.2cm}

The project is partially supported by ShanghaiTech-Megavii Joint Lab, in part by the National Natural Science Foundation of China
(NSFC) under Grants No. 61932020, and supported by the ShanghaiTech-UnitedImaging Joint Lab, Ningbo “2025 S\&T Megaprojects” and Ningbo 3315 Innovation team grant. We also acknowledge the contribution of Weixin Luo and Wen Liu for their insightful comments with regard to the reconstruction-based anomaly detection method.


\bibliographystyle{IEEEbib}
\bibliography{strings}

\begin{thebibliography}{10}

\bibitem{apostolopoulos2017pathological}
Stefanos Apostolopoulos, Sandro De~Zanet, et~al.,
\newblock ``Pathological oct retinal layer segmentation using branch residual
  u-shape networks,''
\newblock in {\em MICCAI}. Springer, 2017, pp. 294--301.

\bibitem{zhao2018uniqueness}
Yitian Zhao, Yalin Zheng, et~al.,
\newblock ``Uniqueness-driven saliency analysis for automated lesion detection
  with applications to retinal diseases,''
\newblock in {\em MICCAI}. Springer, 2018, pp. 109--118.

\bibitem{huang1991optical}
David Huang, Eric~A Swanson, et~al.,
\newblock ``Optical coherence tomography,''
\newblock {\em Science}, vol. 254, no. 5035, pp. 1178--1181, 1991.

\bibitem{krizhevsky2012imagenet}
Alex Krizhevsky, Ilya Sutskever, et~al.,
\newblock ``Imagenet classification with deep convolutional neural networks,''
\newblock in {\em NeurIPS}, 2012, pp. 1097--1105.

\bibitem{lian2018multiview}
Dongze Lian, Lina Hu, et~al.,
\newblock ``Multiview multitask gaze estimation with deep convolutional neural
  networks,''
\newblock {\em IEEE transactions on neural networks and learning systems},
  2018.

\bibitem{lee2017deep}
Cecilia~S Lee, Doug~M Baughman, et~al.,
\newblock ``Deep learning is effective for classifying normal versus
  age-related macular degeneration oct images,''
\newblock {\em Ophthalmology Retina}, vol. 1, no. 4, pp. 322--327, 2017.

\bibitem{zhou2018multi}
Kang Zhou, Zaiwang Gu, et~al.,
\newblock ``Multi-cell multi-task convolutional neural networks for diabetic
  retinopathy grading,''
\newblock in {\em 2018 40th Annual International Conference of the IEEE
  Engineering in Medicine and Biology Society}. IEEE, 2018, pp. 2724--2727.

\bibitem{gu2019net}
Zaiwang Gu, Jun Cheng, et~al.,
\newblock ``Ce-net: Context encoder network for 2d medical image
  segmentation,''
\newblock {\em IEEE transactions on medical imaging}, 2019.

\bibitem{sidibe2017anomaly}
Desire Sidibe, Shrinivasan Sankar, et~al.,
\newblock ``An anomaly detection approach for the identification of dme
  patients using spectral domain optical coherence tomography images,''
\newblock {\em Computer methods and programs in biomedicine}, vol. 139, pp.
  109--117, 2017.

\bibitem{seebock2018unsupervised}
Philipp Seeb{\"o}ck, Sebastian~M Waldstein, et~al.,
\newblock ``Unsupervised identification of disease marker candidates in retinal
  oct imaging data,''
\newblock {\em IEEE TMI}, 2018.

\bibitem{schlegl2017unsupervised}
Thomas Schlegl, Philipp Seeb{\"o}ck, et~al.,
\newblock ``Unsupervised anomaly detection with generative adversarial networks
  to guide marker discovery,''
\newblock in {\em IPMI}. Springer, 2017, pp. 146--157.

\bibitem{goodfellow2014generative}
Ian Goodfellow, Jean Pouget-Abadie, et~al.,
\newblock ``Generative adversarial nets,''
\newblock in {\em Advances in neural information processing systems}, 2014, pp.
  2672--2680.

\bibitem{schlegl2019f}
Thomas Schlegl, Philipp Seeb{\"o}ck, et~al.,
\newblock ``f-anogan: Fast unsupervised anomaly detection with generative
  adversarial networks,''
\newblock {\em Medical Image Analysis}, 2019.

\bibitem{isola2017image}
Phillip Isola, Jun-Yan Zhu, et~al.,
\newblock ``Image-to-image translation with conditional adversarial networks,''
\newblock in {\em CVPR}, 2017, pp. 1125--1134.

\bibitem{akcay2018ganomaly}
Samet Akcay, Amir Atapour-Abarghouei, et~al.,
\newblock ``Ganomaly: Semi-supervised anomaly detection via adversarial
  training,''
\newblock in {\em Asian Conference on Computer Vision}. Springer, 2018, pp.
  622--637.

\bibitem{luo2017revisit}
Weixin Luo, Wen Liu, et~al.,
\newblock ``A revisit of sparse coding based anomaly detection in stacked rnn
  framework,''
\newblock {\em ICCV, Oct}, vol. 1, no. 2, pp. 3, 2017.

\bibitem{luo2019video}
Weixin Luo, Wen Liu, et~al.,
\newblock ``Video anomaly detection with sparse coding inspired deep neural
  networks,''
\newblock {\em IEEE Transactions on Pattern Analysis and Machine Intelligence},
  2019.

\bibitem{zhou2018sc2net}
Joey~Tianyi Zhou, Kai Di, et~al.,
\newblock ``Sc2net: Sparse lstms for sparse coding,''
\newblock in {\em AAAI}, 2018.

\bibitem{zhou2016learning}
Bolei Zhou, Aditya Khosla, et~al.,
\newblock ``Learning deep features for discriminative localization,''
\newblock in {\em CVPR}, 2016, pp. 2921--2929.

\bibitem{kermany2018identifying}
Daniel~S Kermany, Michael Goldbaum, et~al.,
\newblock ``Identifying medical diagnoses and treatable diseases by image-based
  deep learning,''
\newblock {\em Cell}, vol. 172, no. 5, pp. 1122--1131, 2018.

\end{thebibliography}

\end{document}